\documentclass[english, letterpaper, 10pt]{article}

\usepackage[T1]{fontenc}
\usepackage{textcomp}

\usepackage{graphicx}
\usepackage{amssymb,amsmath}
\usepackage{fixltx2e}
\usepackage{fancyhdr}
\usepackage{url}

\usepackage[numbers,sort&compress]{natbib}

\usepackage{dsfont}
\usepackage[top=1in, bottom=1.25in, left=0.875in, right=0.875in]{geometry}
\usepackage{bm}

\usepackage{mathptmx}   % turns on Times New Roman font
\usepackage{titlesec}
\usepackage{titling}

\usepackage{booktabs}
\usepackage{graphicx}
\usepackage{multirow}
\usepackage{subcaption}

\usepackage[english]{babel}
\usepackage[utf8]{inputenc}
\usepackage{amsmath}
\usepackage{amsfonts}
\usepackage{graphicx}
\usepackage{enumitem}
\usepackage[colorinlistoftodos]{todonotes}
\usepackage{abstract}

\usepackage{indentfirst}%Indent in first paragraph

\usepackage{times}%switch to Times New Roman font

\usepackage{titlesec}% formatting of section title
\titleformat{\section}{\bfseries\centering\fontsize{11pt}{13pt}\selectfont}{\thesection.}{4pt}{\uppercase}
\titleformat{\subsection}{\bfseries\fontsize{11pt}{13pt}\selectfont}{\thesubsection}{6pt}{}
\titlespacing*{\section}{0pt}{10pt plus 8pt}{4pt}
\titlespacing*{\subsection}{0pt}{6pt}{3pt}

\usepackage{titling}% formatting of article title
\pretitle{\vspace{-15mm}\begin{center}\bfseries\LARGE}
\posttitle{\end{center}}
\preauthor{\begin{center}\large}
\postauthor{\end{center}\vspace{-15mm}}

\usepackage{etoolbox}
\apptocmd{\thebibliography}{\setlength{\itemsep}{-2pt}}{}{}

\newlength{\myitemsep}
\setlist[itemize]{itemsep=-1pt, topsep=2pt}
\setlist[enumerate]{itemsep=-1pt, topsep=1pt}

\title{Model Selection for Anomaly Detection}
\author{E. Burnaev, P. Erofeev, D. Smolyakov
\\
Institute for Information Transmission Problems (Kharkevich Institute) RAS
}
\date{}

\begin{document}
\pagenumbering{gobble}
\maketitle

\renewcommand{\abstractname}{\vspace{0pt}\fontsize{11pt}{13pt}\selectfont \uppercase{Abstract}\vspace{-4pt}}
\begin{abstract}
\normalsize
Anomaly detection based on one-class classification algorithms is broadly used in many applied domains like image processing (e.g. detection of whether a patient is ``cancerous'' or ``healthy'' from mammography image), network intrusion detection, etc. Performance of an anomaly detection algorithm crucially depends on a kernel, used to measure similarity in a feature space. The standard approaches (e.g. cross-validation) for kernel selection, used in two-class classification problems, can not be used directly due to the specific nature of a data (absence of a second, abnormal, class data). 
In this paper we generalize several kernel selection methods from binary-class case to the case of one-class classification
and perform extensive comparison of these approaches using both synthetic and real-world data.
\vspace{6pt}
\end{abstract}
	
\section{Introduction}
In a one-class classification problem statement, it is assumed that we mainly use data of one, normal, class to build a decision function that describes characteristics of the data. On the other hand data of the other abnormal class are either not used at all or used to refine the obtained data description. The one-class classification is broadly applied to many real application domains namely image processing, network intrusion detection, user verification in computer systems, machine fault detection, etc. In most of the real-world applications for one-class classification, the number of normal data samples is much larger than that of abnormal data samples or abnormal samples are even not available at all \cite{burnaev}. The reason is that collecting normal data is inexpensive and easy in comparison to collecting abnormal data. For example in machine fault detection applications, a normal data can be collected directly under the normal condition of a machine while collecting an abnormal data requires observation of machines until they break. Since the prevalence of one class, in a one-class classification, the boundary decision primarily comes from the dominant class compared to a binary classification, where the data of both classes are used to construct the boundary decision.

One of the most popular approaches for one-class classification is based on support vector models. These models have a nice inherited property allowing to build simple linear decision boundaries in highly non-linear infinite-dimensional spaces thanks to the kernel trick. The most wide-spreadly used kernels are Gaussian kernels. Although these techniques became extremely popular in recent years, there still exists an open question of the kernel bandwidth selection, which, as in binary classification, is crucial for model performance. Despite similarity to the binary classification problems, the standard approaches for bandwidth selection, like cross-validation, can not be used for the one-class classification models due to specific nature of the data (absence of second, abnormal, class data). 

% \begin{wrapfigure}{l}{0.4\textwidth}\label{SVDD_example}
% 	  %\vspace{-15pt}
% 	\includegraphics[width=\linewidth]{SVDD_explanation.png}
% 		  %\vspace{-25pt}
% 	\caption{SVDD illustration.}
% 	  %\vspace{-40pt}	  
% \end{wrapfigure}

In this paper we concentrate on Support Vectors Data Description (SVDD) \cite{tax2004support} which is a one-class 
classification algorithm from the family of support vector models. The main difference from a standard one-class SVM is that in a linear case SVDD builds a sphere surface around the normal data instead of a linear hyperplane, separating the normal data from the origin (in a feature space).
%It is widely used in many applied problems like an outliers detection\cite{hodge2004survey}, an intrusion detection\cite{wang2004anomaly}, a novelty detection\cite{gardner2006one}.
% Main feature of the SVDD is a possibility of using different kernels. 
% It allows to build decision rules in high dimensional spaces without mapping data into them directly. 
% Performance depends strictly on the kernel parameters. 
% In two class applications widely used cross-validation approach. 
% However it is very time consuming and hardly generalizing to one-class problems.
% Despite the wide usage of support-vector-based one-class-classification techniques,

Several approaches for kernel selection were proposed in the literature. Some of them are intended for binary classification but can be generalized to the one-class problem, some are specially designed for one-class classification problem.
For example, in \cite{steinwart2005classification} authors proposed an empirical risk for one-class SVM and proved that under certain conditions it converges to the real risk. Another approach \cite{lukashevich2009using} is based on an idea that a kernel width should be selected in such a way that a fraction of outliers and support vectors in the final model should be close to their corresponding theoretical estimates, depending on parameters of the SVDD algorithm.
In \cite{xiao2014two} authors try to optimize the decision region directly in order to get neither too sparse nor too dense area. All the mentioned approaches imply building and comparing several models w.r.t. some criterion, while in \cite{evangelista2007some} a direct optimization method for kernel parameters is proposed without explicit model building. 

In this paper we propose an original method for SVDD model selection based on empirical risk approximation using oversampling with SMOTE \cite{chawla2002smote} and develop a methodology for SVDD model selection. Finally, we generalize several kernel selection methods from binary-class case to the case of one-class classification
and perform extensive comparison of these approaches using both synthetic and real-world data.

\section{Anomaly Detection using SVDD}

We consider Support Vector Data Description (SVDD) proposed in \cite{tax2004support} for the one-class classification problem. Let us have a dataset $D = \{X_1,\ldots, X_l\},$ $X_i\in \mathbb{R}^n$. Let $\phi_{\gamma}(\cdot)$ be some mapping to a high-dimensional space depending on scalar parameter $\gamma\in \mathbb{R}_{+}$. In this space we solve the following optimization problem:

\begin{equation}
\begin{cases}
R^2 + \frac{1}{\nu l}\sum_{i=1}^l \xi_i \to \min_{R,a}\\
\|\phi_{\gamma}(X_i) - a \|^2 \leqslant R^2 + \xi_i &, i = 1,\ldots, l,\\
\xi_i \geqslant 0, & i = 1, \ldots, l.
\end{cases}\label{eq:svdd}
\end{equation}
Physical meaning of such an optimization problem statement is that we are looking for a minimum volume ball in high-dimensional space defined by $\phi_{\gamma}(\cdot)$, containing the dataset. 
Herewith we allow some points to be outside the boundary of the ball and controll their number by $\nu$. The penalty in (\ref{eq:svdd}) is equivalent to $l_1$ penalty that zeros out some of the parameters. In our case it means that several points will belong exactly to the surface of the ball \cite{tax2004support}. We can re-write \eqref{eq:svdd} in dual form and find that a decision function, predicting whether a point is an anomaly or not, is defined through the kernel $K_{\gamma}(X,X') = \langle\phi_{\gamma}(X),\phi_{\gamma}(X')\rangle$ and has the form $f_{\gamma}(X) = \mathrm{sign}\left(\sum_{i=1}^l\alpha_iK_{\gamma}(X,X_i)-\rho\right)$ for some $\alpha_i\geq0$ and $\rho$. A typical example of the kernel we are using is $K_{\gamma}(X,X') = \exp\left(-\|X-X'\|^2/\gamma\right)$, $\gamma>0$.

\section{Model Selection Techniques}
In this section we give brief descriptions of several model selection approaches for SVDD, which are based on optimization of some risk functions.

\subsection{Support Vectors and Anomalies Fraction Optimization}
Complexity of the model is described by the number of support vectors $X_i\in D$ corresponding to $\alpha_i>0$. In the original paper \cite{tax2004support} authors proved that parameter $\nu$ in (\ref{eq:svdd}) is an upper bound for the fraction of elements in the train data marked as anomaly (outliers) and a lower bound for the fraction of elements used as support vectors. 
Let us define the risk function as follows
\begin{equation}
R^{\textrm{SV}}_{\gamma} = (\nu - f_{\textrm{SV},\gamma})^2,
\end{equation}
where $f_{\textrm{SV},\gamma}$ is a fraction of support vectors in the dataset, defining the current decision function $f_{\gamma}(X)$.
Minimizing $R^{\textrm{SV}}_{\gamma}$ we can balance complexity and classification accuracy on the train set.

\subsection{Empirical Risk}
Another approach would be to reduce the one-class classification to binary classification  \cite{steinwart2005classification}. It can be proved that the following empirical risk converges to a real risk of this binary classification problem
\begin{equation}\label{empiric}
R_{\gamma}^{\textrm{empirical}} = \frac{1}{(1-\nu) \cdot l}\sum_{i=1}^n [f_{\gamma}(X_i) = -1] + 
\frac{1}{\nu}\mathbb{E}_{\mu}[f_{\gamma}(X_i) = 1],
\end{equation}
where $\mathbb{E}_{\mu}[f(X_i) = 1]$ denotes average error of classification on anomaly elements assuming that they have distribution $\mu$. 
This value is estimated using Monte Carlo simulation.
Generally, for modeling $\mu$ it is reasonable to use a least favorable uniform distribution.

\subsection{Risk based on Oversampling}
In empirical risk (\ref{empiric}) the same dataset is used for a model construction and its risk estimation leading to a biased estimates.
% If data set is small it can make some problems. 
That is why for the risk assessment we propose to use sampling based on SMOTE \cite{chawla2002smote}, an oversampling technique, designed for the imbalanced binary classification. We define the risk as
\begin{equation}\label{smote}
R_{\gamma}^{\textrm{SMOTE}} = \frac{1}{(1-\nu) \cdot m}\sum_{i=1}^m \left[f_{\gamma}\left(\tilde{X}_i\right) = -1\right] + 
\frac{1}{\nu}\mathbb{E}_{\mu}[f_{\gamma}(X_i) = 1]
\end{equation}
New examples $\tilde{X}_i$ of a normal data are generated synthetically based on the training set and the SMOTE algorithm. The first summand
in (\ref{smote}) is proportional to the number of synthetic normal elements marked by the classifier as anomalies. The second summand is the same as in (\ref{empiric}).

\subsection{Kernel Matrix Optimization}
Another popular idea is to optimize a kernel matrix directly without building an anomaly detection model.  For example, in \cite{evangelista2007some} a simple statistic was proposed:
\begin{equation}\label{kernel_variance}
R_{\gamma}^{\mathrm{kernel}} = \dfrac{\bar{k}_{\gamma}}{s^2_{\gamma}},
\end{equation}
where $\bar{k}_{\gamma} = \frac{\sum_{i=2}^n\sum_{j=i}^n K_{\gamma}(X_i, X_j)}{(n-1)(n-2)}$ and $s^2_{\gamma} = \frac{\sum_{i=2}^n\sum_{j=i}^n (\bar{k}_{\gamma} - K_{\gamma}(X_i, X_j))^2}{(n-1)(n-2)}
$ are a normalized sum of the kernel matrix elements and a normalized variance of the kernel matrix elements correspondingly.

\subsection{Kernel Polarization}
Another approach for tuning the kernel matrix is a so-called polarization optimization. It was originally proposed for binary classification problems but can be naturally generalized for one class classification problems \cite{steinwart2005classification}. 
In order to calculate the polarization we generate an artificial anomaly data $\{X_{l+1},\ldots,X_{2l}\}$ from a uniform distribution. 
All elements of the initial data set are marked as normal data and finally we have a set of elements $(X_i,y_i)$, $i = 1,\ldots,2l$, where $y_i = 1$ for $i = 1,\ldots,l$ and $y_i = -1$ for $i = l+1,\ldots,2l$. The polarization risk is calculated as 
\cite{baram2005learning}
\begin{equation}
R_{\gamma}^{\mathrm{polar}} = -\sum_{i, j=1}^{2l} y_i \cdot K_{\gamma}(X_i, X_j) \cdot y_j.
\end{equation}

\section{Numerical Experiments}

In this section we describe our settings for the numerical experiments and provide results of kernel selection techniques comparison.

\subsection{Validation Error}
The real risk of anomaly detection consists of two parts: misclassification of points from a normal distribution as anomalies and misclassification of abnormal data as a normal.

For measuring the quality of a decision function we will use sum of these errors on independent validation data set normalized by the number of elements. 
Let $\left\{X^{\mathrm{normal}}_1, \cdots, X^{\mathrm{normal}}_s\right\}$ be the set of normal elements in the validation set and $\left\{X_1^{\mathrm{anomaly}}, \cdots, X_m^{\mathrm{anomaly}}\right\}$ be the set of anomaly elements in the validation set.  Then the  real risk estimate is calculated as follows:
\begin{equation}
R^{\mathrm{val}}_{\gamma} = \frac{1}{s}\sum_{i=1}^s\left[f_{\gamma}\left(X_i^{\mathrm{normal}}\right) = -1\right] 
+ \frac{1}{m}\sum_{j=1}^m\left[f_{\gamma}\left(X_j^{\mathrm{anomaly}}\right) = 1\right].
\end{equation}

The validation risk behavior depends on a dimensionality $n$ of a data. A typical behavior is presented in figure~\ref{fig:validation-risk}. Generally there is a plateau of almost the same validation error values for some range of $\gamma>0$.

\subsection{Behavior of the Risk Functions}
In order to understand a typical behavior of the introduced risk functions we estimate $R^{\mathrm{val}}_{\gamma}$ of the constructed model for different values of $\gamma$ and compare it to the estimates of the risks.
In these experiments we fix the value of $\nu$ to $0.1$, meaning that we have $10\%$ of anomalies.
We generate $100$ points of normal data from a mixture of two $5$-dimensional normal distributions with means at $[1, 1, 1, 1, 1]^{T}$ and $[-1, -1, -1, -1, -1]^{T}$, and identity covariance matrices. Also we generate $100$ anomalies from the uniform distribution on $[-5,5]^{5}$. For the validation a huge dataset from the same distributions is drawn.

\begin{figure}[t!]
%\vskip -0.5in
   \centering
   \begin{subfigure}[b]{0.40\textwidth}
       \includegraphics[width=\textwidth]{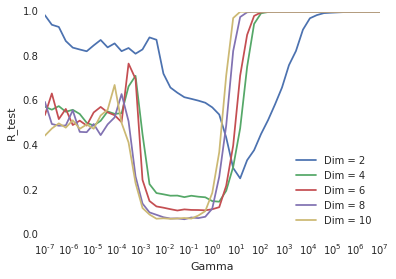}
       \caption{Validation error behavior}
       \label{fig:validation-risk}
   \end{subfigure} ~ \begin{subfigure}[b]{0.40\textwidth}
       \includegraphics[width=\textwidth]{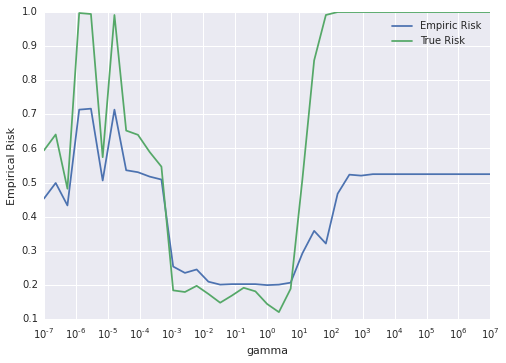}
       \caption{Empirical risk}
       \label{fig:empirical-risk-artificail}
   \end{subfigure}
   
   \begin{subfigure}[b]{0.40\textwidth}
       \includegraphics[width=\textwidth]{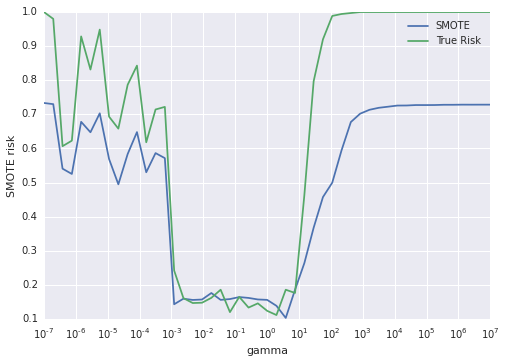}
       \caption{SMOTE risk}
       \label{fig:smote-risk-artificial}
   \end{subfigure} ~ \begin{subfigure}[b]{0.40\textwidth}
       \includegraphics[width=\textwidth]{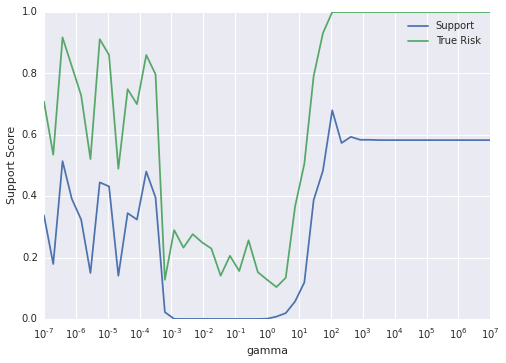}
       \caption{SV risk}
       \label{fig:support-risk-artificial}
   \end{subfigure}
   
   \begin{subfigure}[b]{0.40\textwidth}
       \includegraphics[width=\textwidth]{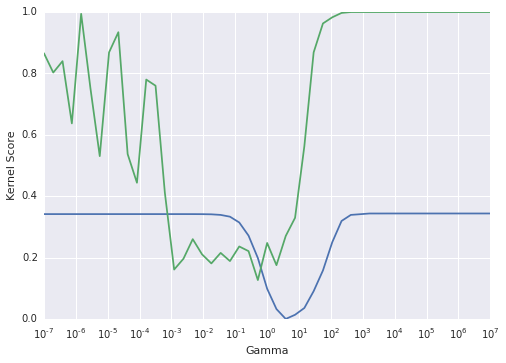}
       \caption{Kernel risk}
       \label{fig:kernel-risk-artificial}
   \end{subfigure} ~ \begin{subfigure}[b]{0.40\textwidth}
       \includegraphics[width=\textwidth]{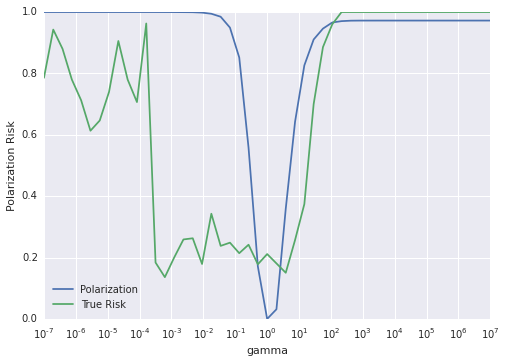}
       \caption{Polarization risk}
       \label{fig:polarization-risk-artificial}
   \end{subfigure}
   \caption{Dependency of a risk function and a validation error on $\gamma$}
\end{figure}

Empirical risk (see figure~\ref{fig:empirical-risk-artificail}) looks very similar to the validation error, ranges of plateaus mostly coincide. Risk based on SMOTE oversampling (see figure~\ref{fig:smote-risk-artificial}) looks even more similar to the real validation error. But in either cases choosing the minimizer of the risk as an optimal value for $\gamma$ we obtain similar results. In case of support vectors based risk (see figure~\ref{fig:support-risk-artificial}) the plateau of the risk is much wider than those of the validation error. The kernel risk (see figure~\ref{fig:kernel-risk-artificial}) behaves very smoothly with a single minimum as it was shown in \cite{evangelista2007some} but this mimimum is biased with respect to the optimal value of $\gamma$. Finally, the polarization risk (see figure~\ref{fig:polarization-risk-artificial}) also gives a single minimum and the optimum value belongs to the plateau of the validation error.

\subsection{Distribution of Anomalies}
\begin{figure}[t!]
%\vskip -0.5in
   \centering
   \begin{subfigure}[b]{0.40\textwidth}
       \includegraphics[width=\textwidth]{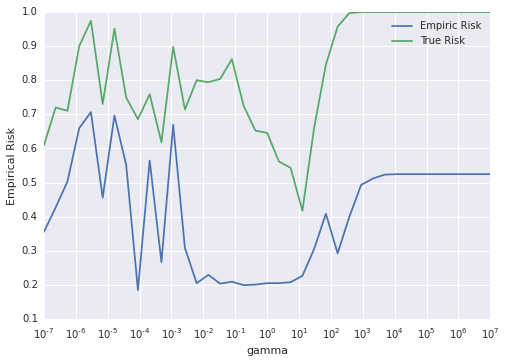}
       \caption{Empiric risk}
       \label{fig:gull}
   \end{subfigure} ~ \begin{subfigure}[b]{0.40\textwidth}
       \includegraphics[width=\textwidth]{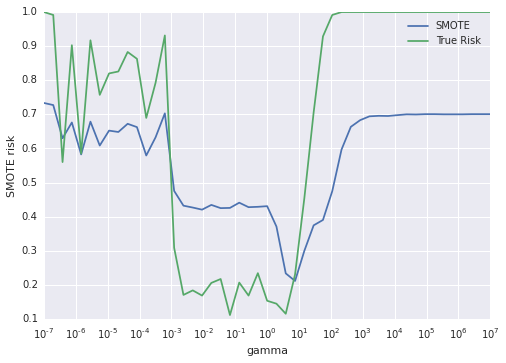}
       \caption{SMOTE risk}
       \label{fig:tiger}
   \end{subfigure}
   
   \begin{subfigure}[b]{0.40\textwidth}
       \includegraphics[width=\textwidth]{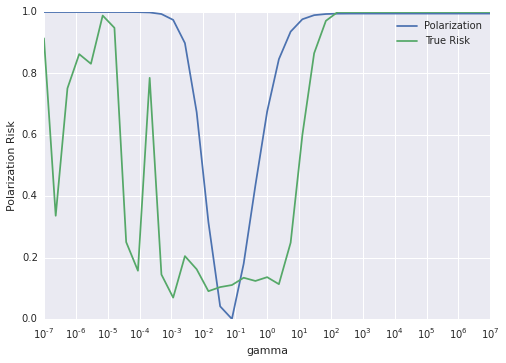}
       \caption{Polarization risk}
       \label{fig:gull}
   \end{subfigure} ~ \begin{subfigure}[b]{0.40\textwidth}
       \includegraphics[width=\textwidth]{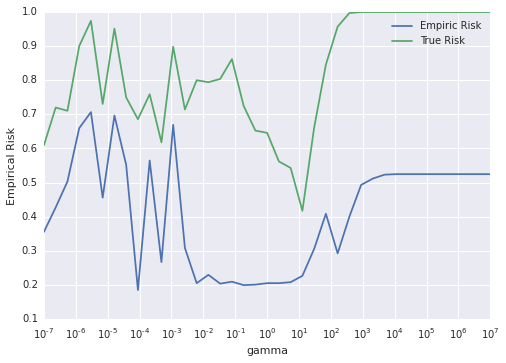}
       \caption{Empiric risk}
       \label{fig:gull}
   \end{subfigure}
   
   \begin{subfigure}[b]{0.40\textwidth}
       \includegraphics[width=\textwidth]{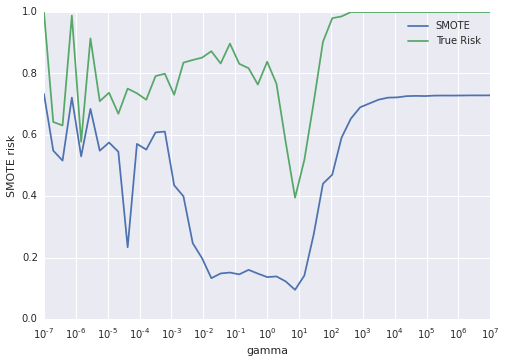}
       \caption{SMOTE risk}
       \label{fig:tiger}
   \end{subfigure} ~ \begin{subfigure}[b]{0.40\textwidth}
       \includegraphics[width=\textwidth]{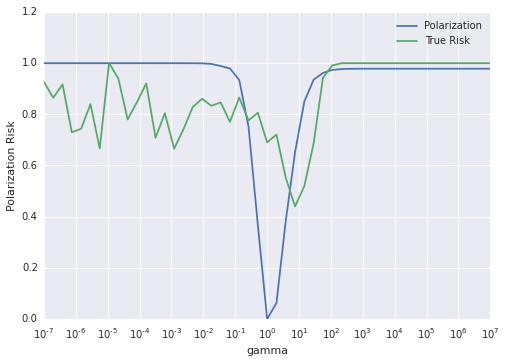}
       \caption{Polarization risk}
       \label{fig:gull}
   \end{subfigure}
  \caption{Dependency of a risk function and a validation error on $\gamma$ for a non-uniform anomaly distribution}
  \label{normal-anomaly}
\end{figure}

We use a uniform distribution of anomalies
when calculating the proposed risk functions.
It is natural to test how these risk functions depend on the distribution of anomalies. We use a normal distribution with variance, comparable to the range of the normal data, to generate anomalies in the learning data sample. At the same time when estimating risks we use uniformly distributed synthetic anomalies. From figure \ref{normal-anomaly} we see that it is still possible to use uniformly generated synthetic anomalies even if true anomalies are generated from another distribution; and we can select a reasonable $\gamma$ as the largest value from the plateau of the corresponding risk function.

\subsection{Real World Data}
Real world data were taken from 
\url{http://homepage.tudelft.nl/n9d04/occ/index.html}.
These data sets are generated from multi class classification problems like ``Iris Dataset'', ``Sonar'', etc. For these problems we considered each class as normal and added anomalies
from a uniform distribution to get a one-class classification problem. Borders of the uniform 
distribution were taken as double borders of the normal class. Number of anomalies were equal to $5\%$, $10\%$ and $15\%$ of a normal class size. Finally we constructed $96$ datasets with smallest value of $n$ equal to $5$ and biggest value of $n$ equal to $1909$. For some of the datasets $l\ll n$ (datasets, obtained from ``Leukemia'' and ``Colon'' classification problems). For majority of datasets $l\leq 1000$, but also there are several big datasets like those, obtained from ``Spam Base'' ($l = 4600$) 
and ``Concordia'' ($l=4000$) classification problems. Also there are data sets with a very small 
$l$ like that obtained from ``Leukemia'' ($l = 72$) classification problem.

To compare model selection methods on real data we use Dolan-More curves \cite{dolan-more} which are built in the following way. Let $\{R_1, \ldots, R_K\}$ be the set of considered model selection methods, $\{D_1, \ldots, D_M\}$ be the set of tasks (datasets), $q_{ti}$ be the quality of the method $i$ on the dataset $t$. For each method $i$ we introduce $p_i(\beta)$, a fraction of datasets, on which the method $i$ is worse than the best one not more than $\beta$ times:
$$
p_i(\beta) = \frac{1}{T} \left| \left\{t: q_{ti} \geq \frac{1}{\beta} \max\limits_{i} q_{ti} \right\}\right|, \ \beta \geq 1.
$$
For example, $p_i(1)$ is a fraction of datasets where the method~$i$ is the best. A graph of $p_i(\beta)$ is called Dolan-More curve for the method $i$. This definition implies that the higher the curve, the better the method.
Note that Dolan-More curve for a particular method depends on other methods considered in comparison. 

\begin{figure}[t!]
\begin{center}
\includegraphics[resolution=900, scale=0.6]{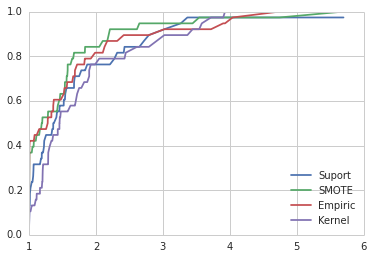}
\end{center}
  \caption{Dolan-More curves for comparison of model selection methods on real data}
  \label{dolan2002benchmarking}
\end{figure}
From figure \ref{dolan2002benchmarking} we can see that perfomances of the SMOTE risk and of the Empirical risk are quite similar,  but the curve for the SMOTE risk is slightly higher than that for the Empirical risk. The kernel risk and the Support Vectors risk provides worse perfomance.

\section{Conclusions}
We considered Support Vectors Data Description as a one-class 
classification algorithm; generalized several model selection methods from binary classification for SVDD model selection
and performed extensive comparison of these approaches using synthetically generated data and real-world data sets.

\textbf{Acknowledgement}: The research was conducted in IITP RAS and supported solely by the Russian Science Foundation grant (project 14-50-00150).

\renewcommand\refname{\uppercase{References}}


\begin{thebibliography}{10}

\bibitem{tax2004support}
David~MJ Tax and Robert~PW Duin.
\newblock Support vector data description.
\newblock {\em Machine learning}, 54(1):45--66, 2004.

\bibitem{hodge2004survey}
Victoria~J Hodge and Jim Austin.
\newblock A survey of outlier detection methodologies.
\newblock {\em Artificial Intelligence Review}, 22(2):85--126, 2004.

\bibitem{wang2004anomaly}
Yanxin Wang, Johnny Wong, and Andrew Miner.
\newblock Anomaly intrusion detection using one class svm.
\newblock In {\em Information Assurance Workshop, 2004. Proceedings from the
  Fifth Annual IEEE SMC}, pages 358--364. IEEE, 2004.

\bibitem{gardner2006one}
Andrew~B Gardner, Abba~M Krieger, George Vachtsevanos, and Brian Litt.
\newblock One-class novelty detection for seizure analysis from intracranial
  eeg.
\newblock {\em The Journal of Machine Learning Research}, 7:1025--1044, 2006.

\bibitem{steinwart2005classification}
Ingo Steinwart, Don~R Hush, and Clint Scovel.
\newblock A classification framework for anomaly detection.
\newblock In {\em Journal of Machine Learning Research}, pages 211--232, 2005.

\bibitem{lukashevich2009using}
Hanna Lukashevich, Stefanie Nowak, and Peter Dunker.
\newblock Using one-class svm outliers detection for verification of
  collaboratively tagged image training sets.
\newblock In {\em Multimedia and Expo, 2009. ICME 2009. IEEE International
  Conference on}, pages 682--685. IEEE, 2009.

\bibitem{xiao2014two}
Yingchao Xiao, Huangang Wang, Lin Zhang, and Wenli Xu.
\newblock Two methods of selecting gaussian kernel parameters for one-class svm
  and their application to fault detection.
\newblock {\em Knowledge-Based Systems}, 59:75--84, 2014.

\bibitem{evangelista2007some}
Paul~F Evangelista, Mark~J Embrechts, and Boleslaw~K Szymanski.
\newblock Some properties of the gaussian kernel for one class learning.
\newblock In {\em Artificial Neural Networks--ICANN 2007}, pages 269--278.
  Springer, 2007.

\bibitem{chawla2002smote}
Nitesh~V Chawla, Kevin~W Bowyer, Lawrence~O Hall, and W~Philip Kegelmeyer.
\newblock Smote: synthetic minority over-sampling technique.
\newblock {\em Journal of artificial intelligence research}, 16(1):321--357,
  2002.

\bibitem{wang2009learning}
Ting-hua Wang, Hou-kuan Huang, Sheng-feng Tian, and Dayong Deng.
\newblock Learning general gaussian kernels by optimizing kernel polarization.
\newblock {\em Chinese Journal of Electronics}, 18(2):265--269, 2009.

\end{thebibliography}

\begin{thebibliography}{10}

\bibitem{burnaev}
E. Burnaev, P. Erofeev, A. Papanov. ``Influence of Resampling on Accuracy of Imbalanced Classification'', Proceedings of the ICMV-2015 conference (2015) 

\bibitem{tax2004support}
D. Tax, R. Duin.
\newblock ``Support vector data description'',
\newblock Machine learning, 54(1), p. 45--66 (2004)

\bibitem{steinwart2005classification}
I. Steinwart, D. Hush, C. Scovel.
\newblock ``A classification framework for anomaly detection'',
\newblock {Journal of Machine Learning Research}, p. 211--232 (2005)

\bibitem{lukashevich2009using}
H. Lukashevich, S. Nowak, P. Dunker.
\newblock ``Using one-class svm outliers detection for verification of
  collaboratively tagged image training sets'',
\newblock {IEEE International
  Conference on Multimedia and Expo ICME-2009}, p. 682--685, IEEE (2009)

\bibitem{xiao2014two}
Y. Xiao, H. Wang, L. Zhang, W. Xu.
\newblock ``Two methods of selecting gaussian kernel parameters for one-class svm and their application to fault detection'',
\newblock {Knowledge-Based Systems}, vol. 59, p. 75--84 (2014)


\bibitem{evangelista2007some}
P. Evangelista, M. Embrechts, B. Szymanski.
\newblock ``Some properties of the gaussian kernel for one class learning'',
\newblock {Artificial Neural Networks--ICANN 2007}, p. 269--278, Springer (2007)

\bibitem{chawla2002smote}
N. Chawla, K. Bowyer, L. Hall, W. Kegelmeyer.
\newblock ``Smote: synthetic minority over-sampling technique'',
\newblock {Journal of artificial intelligence research}, 16(1), p. 321--357 (2002)


\bibitem{hodge2004survey}
V. Hodge, J. Austin.
\newblock ``A survey of outlier detection methodologies'',
\newblock {Artificial Intelligence Review}, 22(2), p.85--126 (2004)

\bibitem{wang2004anomaly}
Y. Wang, J. Wong, A. Miner.
\newblock ``Anomaly intrusion detection using one class svm'',
\newblock {Proceedings from the
  Fifth Annual IEEE SMC Information Assurance Workshop 2004}, p. 358--364, IEEE (2004)

\bibitem{gardner2006one}
A. Gardner, A. Krieger, G. Vachtsevanos, B. Litt.
\newblock ``One-class novelty detection for seizure analysis from intracranial EEG'',
\newblock {The Journal of Machine Learning Research}, vol. 7, p. 1025--1044 (2006)

\bibitem{baram2005learning}
Y. Baram.
\newblock ``Learning by Kernel Polarization'',
\newblock {Neural Computation}, 17(6), p. 1264-1275 (2005)

\bibitem{dolan-more}
E. Dolan, J. More.
``Benchmarking Optimization Software With Performance Profiles'', Mathematical Programming, 91(2), p.~201-213 (2002)

\end{thebibliography}
\end{document}